\documentclass{article}
\usepackage{spconf,amsmath,graphicx}
\usepackage{multirow}
\usepackage{subcaption}

\title{Revisiting Role of Autoencoders in Adversarial Settings}
\name{Byeong Cheon Kim, Jung Uk Kim, Hakmin Lee, and Yong Man Ro*\thanks{*Corresponding author (ymro@ee.kaist.ac.kr). This work was conducted by Center for Applied Research in Artificial Intelligence (CARAI) grant funded by DAPA and ADD (UD190031RD).}}
\address{Image and Video Systems Lab, School of Electrical Engineering, KAIST, South Korea}

\begin{document}
%
\maketitle
\begin{abstract}
To combat against adversarial attacks, autoencoder structure is widely used to perform denoising which is regarded as gradient masking. In this paper, we revisit the role of autoencoders in adversarial settings. Through the comprehensive experimental results and analysis, this paper presents the inherent property of adversarial robustness in the autoencoders. We also found that autoencoders may use robust features that cause inherent adversarial robustness. We believe that our discovery of the adversarial robustness of the autoencoders can provide clues to the future research and applications for adversarial defense. 
\end{abstract}
\vspace{-0.3cm}
\section{Introduction}
\label{sec:1}
Recent advances in deep neural networks (DNNs) have shown remarkable performance improvement in image classification tasks. Despite their astonishing performance, the DNNs are vulnerable to images with imperceptible perturbations~\cite{szegedy2013intriguing, goodfellow2014explaining}. Such perturbed images are called adversarial examples. This adversarial vulnerability can raise severe issues directly related to safety in many applications such as autonomous driving and malware detection. For that reason, the researches related to adversarial robustness are essential to maintain classification performance against these malicious attacks.

To alleviate this adversarial vulnerability, additional modules are actively utilized to complement DNNs vulnerability. Among these methods, autoencoder structures are widely used as additional modules, and they are generally adopted to add denoising functions~\cite{sahay2019combatting, hwang2019puvae, liao2018defense, chen2018comparative}. However, denoising functions depend on gradient masking which makes DNNs gradients contain less information in generating adversarial examples. In recent years, many researches have been successful in attacking gradient masking-based defenses~\cite{carlini2016defensive, athalye2018obfuscated, uesato2018adversarial}. As a result, for precise evaluation, the additional modules should not have gradient masking effect. However, research on autoencoders that do not lead to gradient masking is insufficient inspite of wide usage of autoencoders for adversarial defense.

Therefore, in this paper, we examine the autoencoder in the adversarial robustness point of view. To analyze the robustness of the autoencoder structure in adversarial settings, we perform extensive experiments based on the autoencoder to be directly utilized as additional modules. We find that autoencoders have inherent adversarial robustness in various environments. In detail, using a simple autoencoder, classification performance is improved in both white and black-box environments. Besides, comparing with other gradient masking defense methods, we show that the performance is not based on the gradient masking effect. Moreover, we show that the features used in autoencoders are themselves robust to adversarial attacks. To the best of our knowledge, this is the first work to analyze the adversarial robustness nature of the autoencoder in adversarial settings from various angles. Our analysis can be hints for the cause of existing adversarial defense methods using autoencoders.

The main contributions of our research can be summarized as follows:
\begin{itemize}
	\vspace{-0.3cm}\item{We verify that autoencoders are adversarially robust, not depending on gradient masking effects through massive experimental results.}
	\vspace{-0.3cm}\item{We analyze that the cause of the adversarial robustness of autoencoders may derive from robust features. Moreover, our experimental results show that the robust features can be the same cause of the robustness of adversarial training. Understanding the features used in autoencoders can present the researchers to get an insight into adversarial robustness.}
\end{itemize}
\vspace{-0.5cm}
\section{AUTOENCODER-CLASSIFIER STRUCTURE TO BE ANALYZED AND EXPERIMENTAL SETUP}
\label{sec:format}

\begin{figure}[t]
	\begin{minipage}[b]{1.0\linewidth}
		\centering
		\centerline{\includegraphics[width=8.6cm]{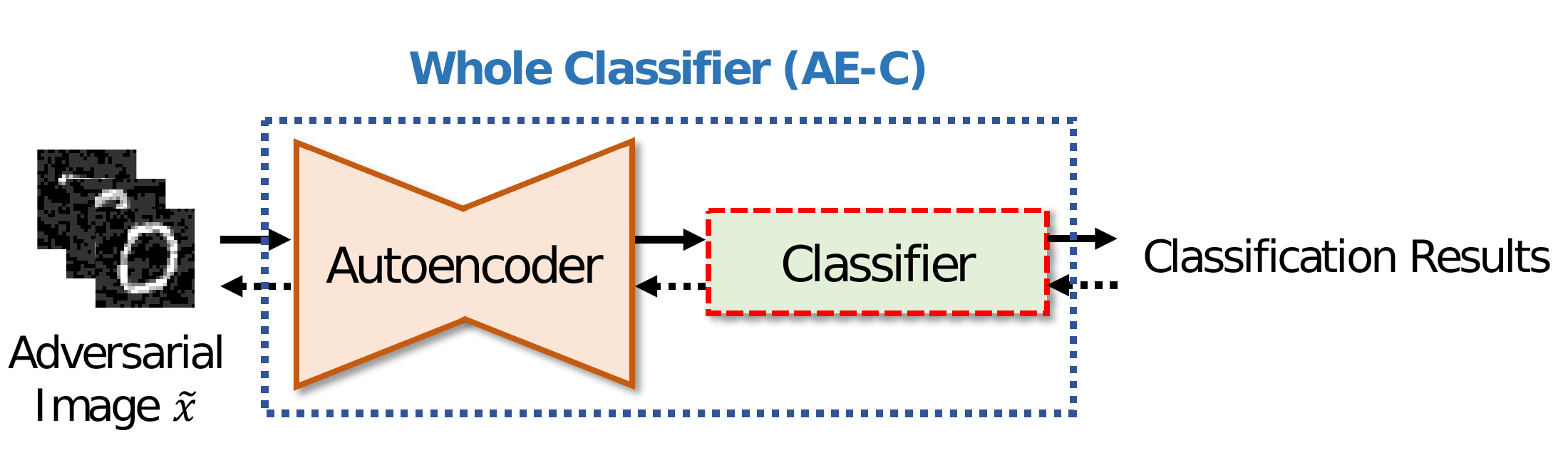}}
	\end{minipage}
	\vspace{-0.3cm}
	\caption{Framework of Whole Classifier (AE-C). Dotted arrow indicates gradient flow of gradient-based attack.}
	\label{fig:1}
\end{figure}

To investigate the adversarial robustness of the autoencoder, we attached an autoencoder in front of the classifier. With this widely used structure in the previous methods~\cite{sahay2019combatting, hwang2019puvae, liao2018defense, chen2018comparative}, we tried to look into the adversarial robustness in various angles. We used pretrained autoencoders using the L2 loss as the objective function. The whole network, namely the whole classifier, AE-C, is shown in Fig.~\ref{fig:1}.  In detail, the encoder and decoder consist of 3 convolution layers with 2 pooling layers and 3 convolution layers with 2 transposed layers, respectively. For a fair comparison, all the backpropagations needed to generate adversarial examples are computed through the whole classifier as represented by the dotted arrows in Fig.~\ref{fig:1}. We use two widely studied datasets MNIST~\cite{lecun1998gradient} and CIFAR10~\cite{krizhevsky2009learning} with two publicly available networks LeNet5~\cite{lecun1998gradient} and VGG16~\cite{simonyan2014very} as vanilla classifiers.
\vspace{-0.3cm}
\section{Experimental Results}
\subsection{Adversarial Robustness of AE-C}
\label{sec:pagestyle}
\subsubsection{Adversarial accuracy against white-box attacks}

\begin{table}[t]
	\centering
		\caption{Accuracy (\%) on adversarial attacks with different perturbations $\varepsilon$. C indicates classifier alone. $\varepsilon$ is the value of when the image is normalized from 0 to 255. We set the confidence c=0 for CW attack.}
	\label{tab:1}
	\begin{tabular}{cccc|ccc}
		& \multicolumn{3}{|c|}{MNIST}          & \multicolumn{3}{c}{CIFAR10}                 \\ \hline
		\multicolumn{1}{c|}{Attack}                & $\varepsilon$ & C     & AE-C           & $\varepsilon$ & C              & AE-C           \\ \hline
		\multicolumn{1}{c|}{$\times$}                     & 0         & 98.23 & \textbf{98.33} & 0         & \textbf{89.67} & 88.32          \\
		\multicolumn{1}{c|}{\multirow{2}{*}{FGSM}} & 20       & 69.94 & \textbf{88.98} & 5         & 24.51          & \textbf{30.47} \\
		\multicolumn{1}{c|}{}                      & 80       & 2.16  & \textbf{50.24} & 10        & 15.24          & \textbf{21.02} \\
		\multicolumn{1}{c|}{\multirow{2}{*}{PGD}}  & 20        & 30.63  & \textbf{71.24} & 5       & 2.58           & \textbf{5.14}  \\
		\multicolumn{1}{c|}{}                      & 80       & 0.00  & \textbf{2.45}  & 10       & 0.19           & \textbf{0.55}  \\
		\multicolumn{1}{c|}{CW}                    & c=0         & 8.50   & \textbf{9.40}   & c=0         & 9.70            & \textbf{10.80}  \\
		\multicolumn{1}{c|}{DF}                    & -         & 0.88   & \textbf{37.67}    & -         & 8.20            & \textbf{9.40}  
	\end{tabular}
\vspace{-0.3cm}
\end{table}

\begin{table}[]
	\centering
		\caption{Accuracy (\%) on black-box adversarial attacks with different perturbations $\varepsilon$. C indicates classifier alone. $\varepsilon$ is the value of when the image is normalized from 0 to 255. (a) Totally new classifiers were used as substitute models. (b) Same structured classifiers were used as substitute models.}
	\begin{subtable}{.48\textwidth}\centering
		\begin{tabular}{c|ccc|ccc}
			
			& \multicolumn{3}{c|}{MNIST}          & \multicolumn{3}{c}{CIFAR10}        \\ \hline
			\multicolumn{1}{c|}{Attack}                & $\varepsilon$ & C     & AE-C           & $\varepsilon$ & C     & AE-C           \\ \hline
			\multicolumn{1}{c|}{\multirow{2}{*}{FGSM}} & 20        & 96.53 & \textbf{97.07} & 5         & 62.07 & \textbf{69.49} \\
			\multicolumn{1}{c|}{}                      & 80       & 69.96 & \textbf{72.95} & 10        & 47.47 & \textbf{57.03} \\
			\multicolumn{1}{c|}{\multirow{2}{*}{PGD}}  & 20        & 96.52 & \textbf{97.19} & 5         & 62.07 & \textbf{69.49} \\
			\multicolumn{1}{c|}{}                      & 80        & 65.64 & \textbf{81.39} & 10        & 28.56 & \textbf{49.45}
		\end{tabular}
		\caption{}
		\label{tab:2.a}
	\end{subtable}
	\begin{subtable}{.48\textwidth}\centering
		\begin{tabular}{c|ccc|ccc}
			& \multicolumn{3}{c|}{MNIST}          & \multicolumn{3}{c}{CIFAR10}        \\ \hline
			\multicolumn{1}{c|}{Attack}                & $\varepsilon$ & C     & AE-C           & $\varepsilon$ & C     & AE-C           \\ \hline
			\multicolumn{1}{c|}{\multirow{2}{*}{FGSM}} & 20        & 92.46 & \textbf{96.24} & 5         & 49.62 & \textbf{61.63} \\
			\multicolumn{1}{c|}{}                      & 80        & 32.03 & \textbf{73.94} & 10        & 28.34 & \textbf{44.36} \\
			\multicolumn{1}{c|}{\multirow{2}{*}{PGD}}  & 20        & 88.20 & \textbf{95.60} & 5         & 42.80 & \textbf{66.33} \\
			\multicolumn{1}{c|}{}                      & 80        & 1.11  & \textbf{71.90} & 10        & 18.28 & \textbf{44.91}
		\end{tabular}
		\caption{}
		\label{tab:2.b}
	\end{subtable}
	\label{tab:2}
	\vspace{-0.8cm}
	
\end{table}

We investigated the adversarial robustness of the framework of Fig.~\ref{fig:1} ($\mathit{i.e.}$ AE-C) in the white-box environment. We applied four adversarial attack methods (FGSM \cite{goodfellow2014explaining}, PGD \cite{madry2017towards}, Carlini \& Wagner (CW) \cite{carlini2017towards}, and DeepFool (DF) \cite{moosavi2016deepfool}). FGSM updates images only once to maximize loss function using gradients and PGD iteratively conducts FGSM. CW optimizes objective function to change predictions with small perturbations. DF iteratively updates inputs with linear approximation. The experimental results are shown in Table~\ref{tab:1}. Following the Table~\ref{tab:1}, adversarial accuracy was increased against various attacks with different size of perturbations. More specifically, in the MNIST experiments, classification accuracy drastically reduces with increasing size of perturbations without an autoencoder. On the other hand, the AE-C maintains the classification accuracy, relatively well. Similar tendencies were shown in stronger attacks such as DF and CW.


However, improved accuracy against white-box attacks does not mean adversarial robustness. Masking the input gradients to impede creating adversarial examples can also improve white-box adversarial accuracy. Defenses using gradient masking is not desired because those defenses are known to be easily destroyed~\cite{carlini2016defensive, athalye2018obfuscated, uesato2018adversarial}. Verification that adversarial accuracy improvement of the AE-C is not caused by the gradient masking effect is conducted in Section~\ref{sec:3.2}.\vspace{-0.2cm}

\subsubsection{Is this caused by gradient masking effects?}
\label{sec:3.2}

\begin{figure*}[t]
	\begin{subfigure}{.25\textwidth}
		\centering
		\includegraphics[width=1\linewidth]{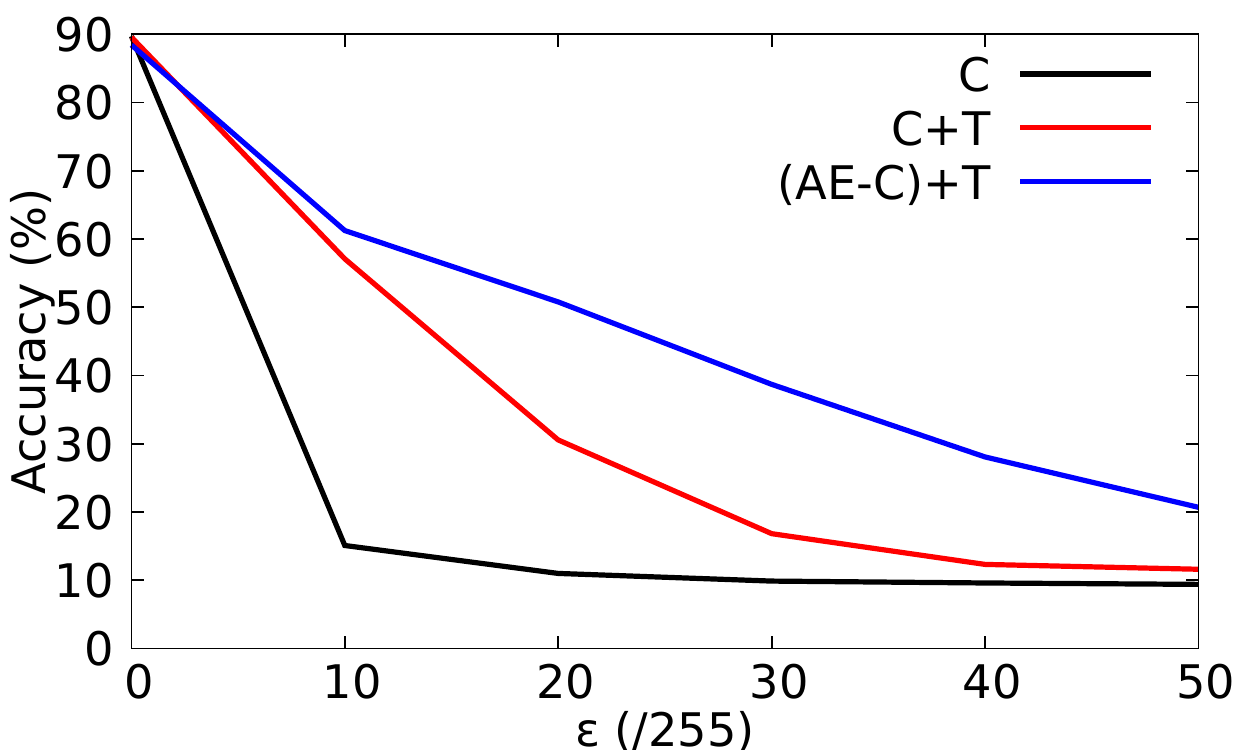}  
		\caption{}
		\label{fig:2a}
	\end{subfigure}\hfill
	\begin{subfigure}{.25\textwidth}
		\centering
		\includegraphics[width=1\linewidth]{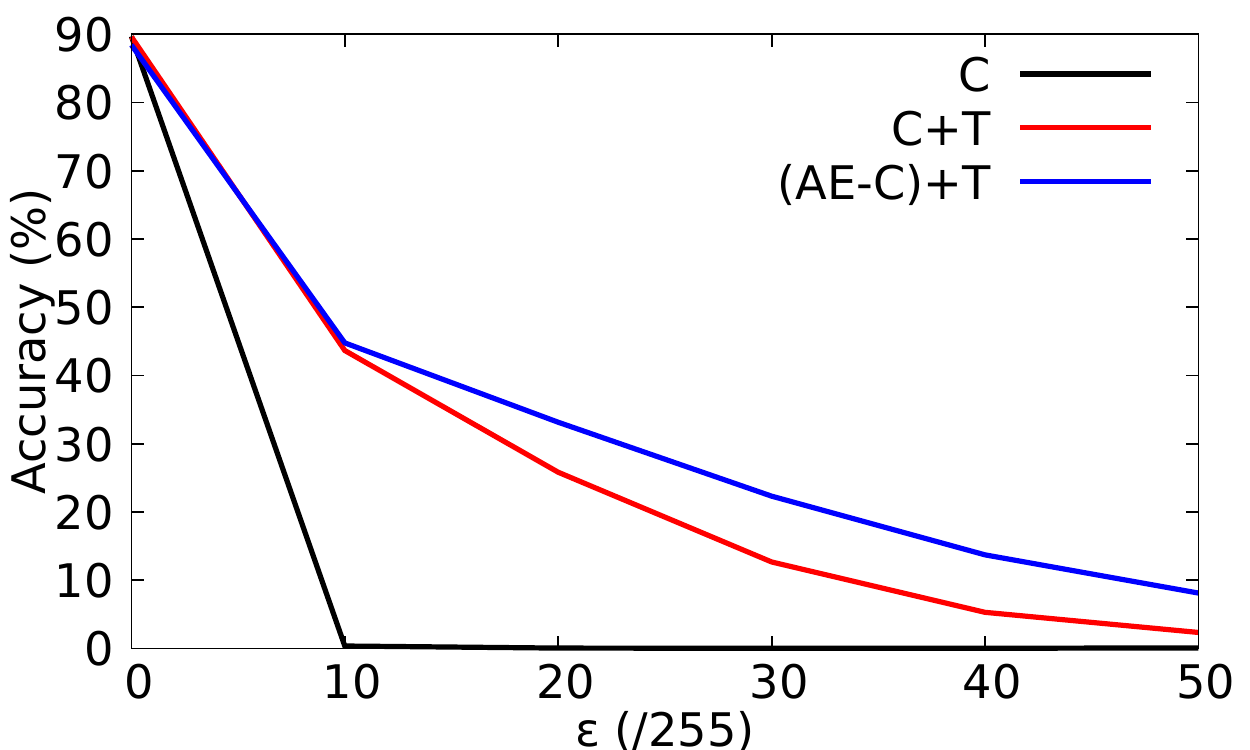}  
		\caption{}
		\label{fig:2b}
	\end{subfigure}\hfill
	\begin{subfigure}{.25\textwidth}
		\centering
		\includegraphics[width=1\linewidth]{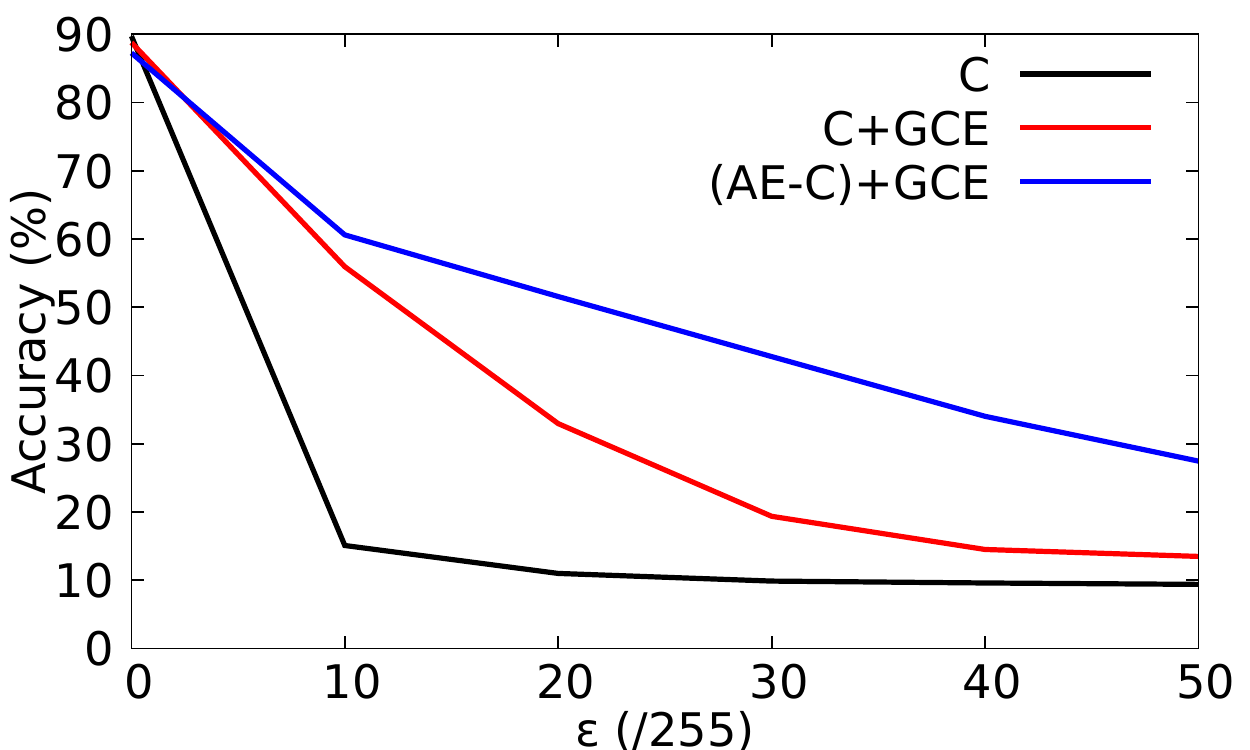}  
		\caption{}
		\label{fig:2c}
	\end{subfigure}\hfill
	\begin{subfigure}{.25\textwidth}
		\centering
		\includegraphics[width=1\linewidth]{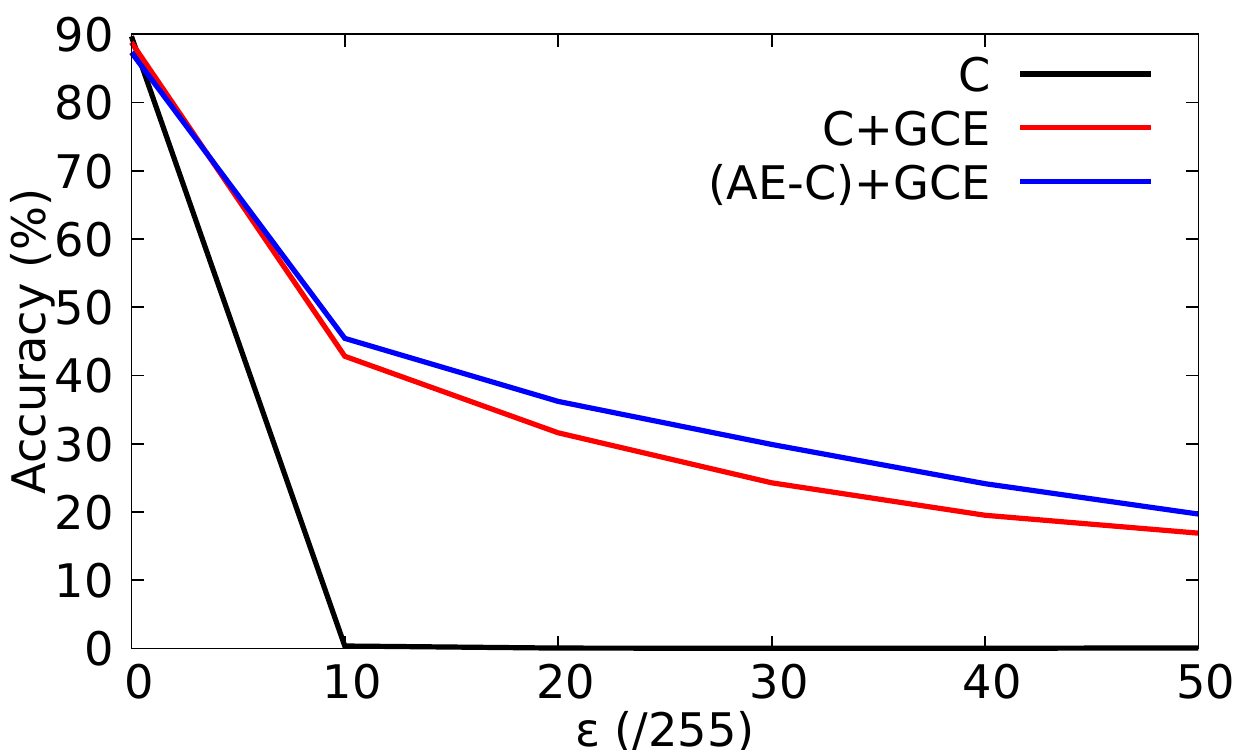}  
		\caption{}
		\label{fig:2d}
	\end{subfigure}
	\begin{subfigure}{.25\textwidth}
		\centering
		\includegraphics[width=1\linewidth]{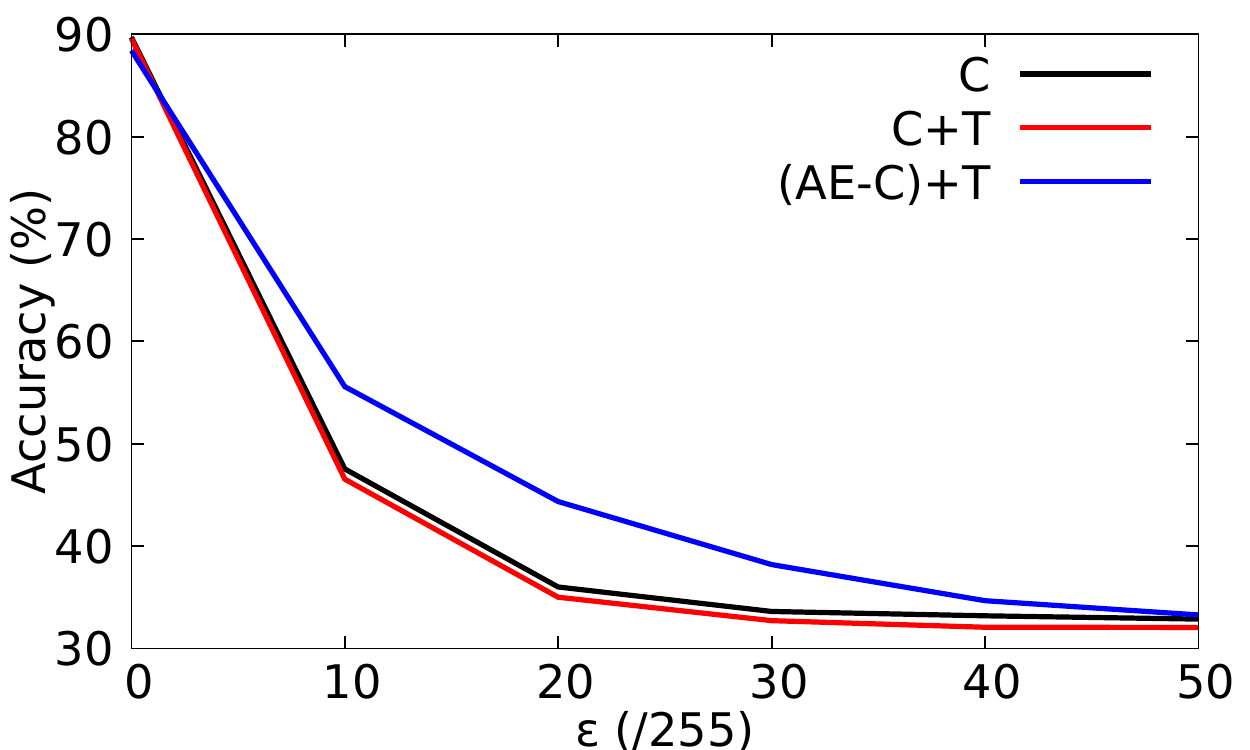}  
		\caption{}
		\label{fig:2e}
	\end{subfigure}\hfill
	\begin{subfigure}{.25\textwidth}
		\centering
		\includegraphics[width=1\linewidth]{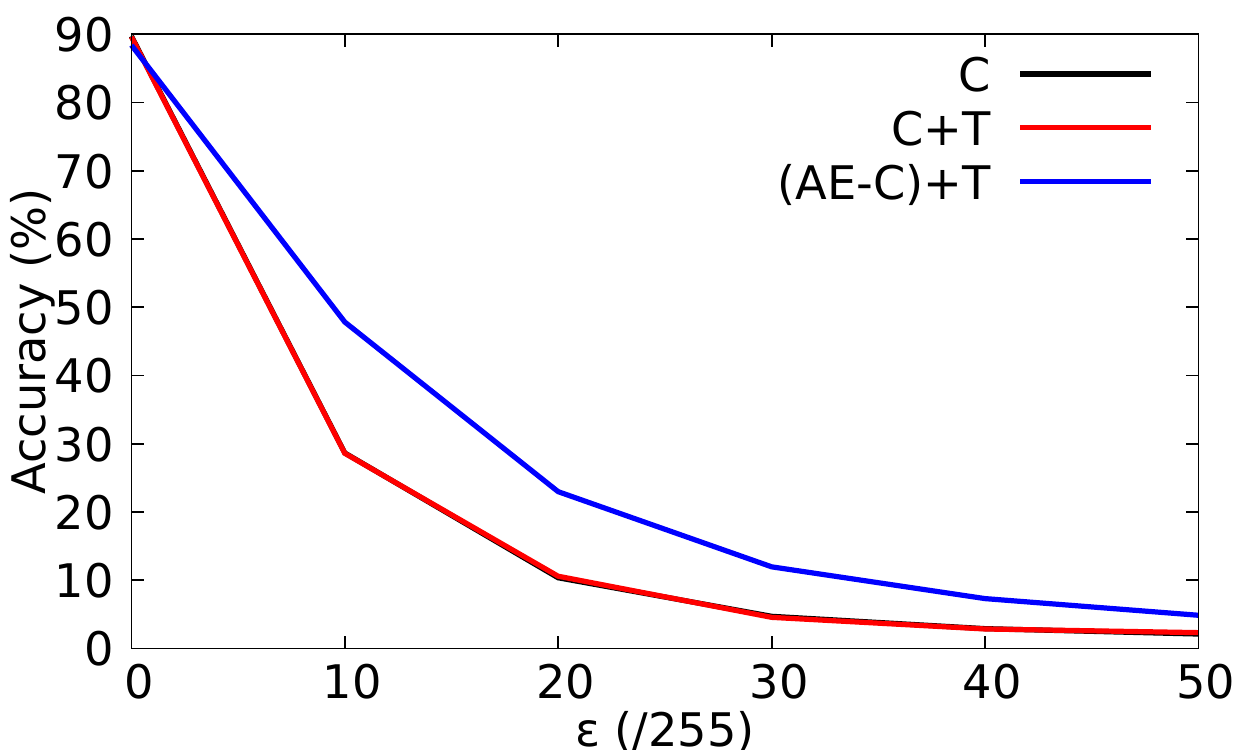}  
		\caption{}
		\label{fig:2f}
	\end{subfigure}\hfill
	\begin{subfigure}{.25\textwidth}
		\centering
		\includegraphics[width=1\linewidth]{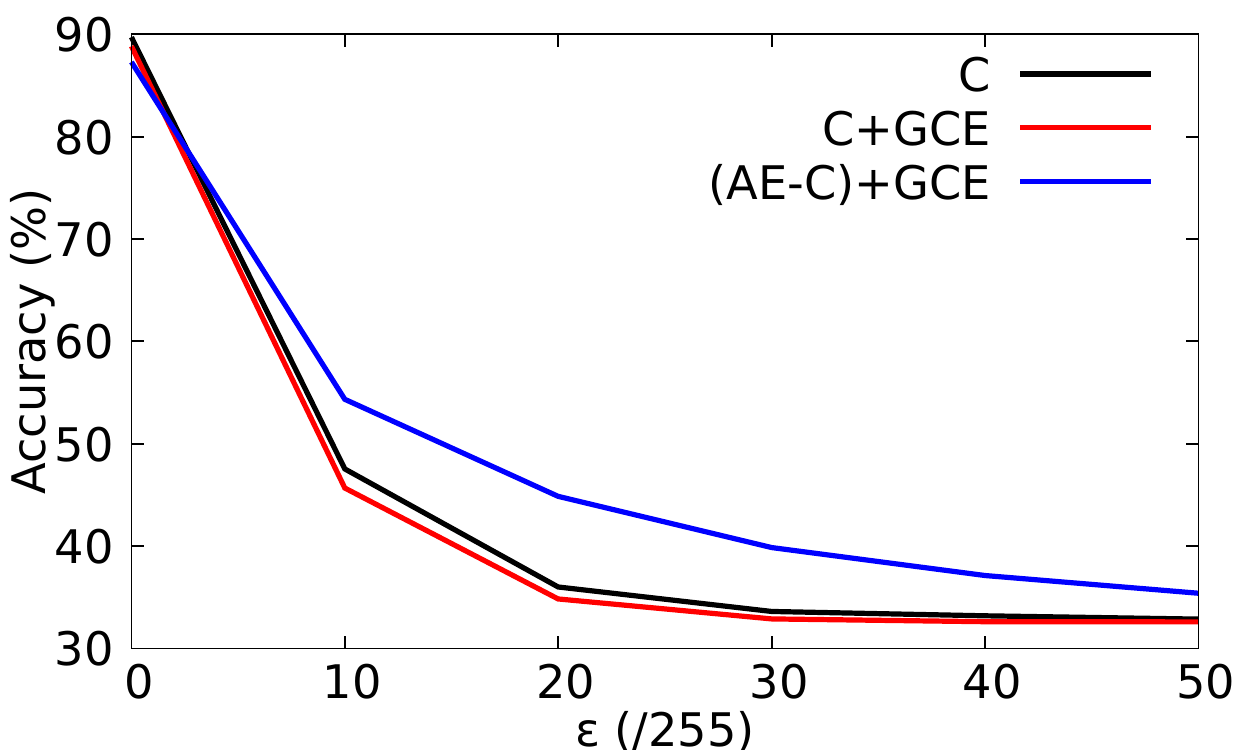}  
		\caption{}
		\label{fig:2g}
	\end{subfigure}\hfill
	\begin{subfigure}{.25\textwidth}
		\centering
		\includegraphics[width=1\linewidth]{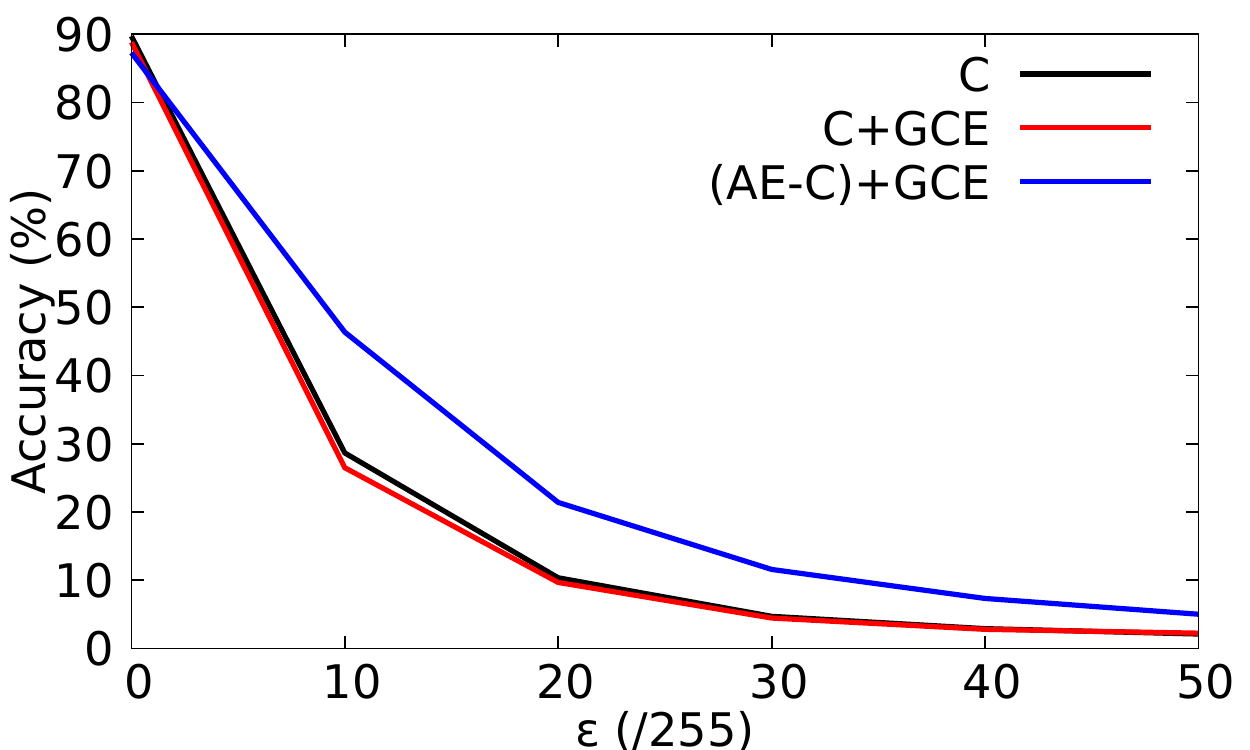}  
		\caption{}
		\label{fig:2h}
	\end{subfigure}
	\vspace{-0.4cm}
	\caption{Comparison of using an autoencoder with other defense methods on CIFAR10 dataset. Experiments in first row and second row are done in white and black-box environments, respectively. C+T and (AE-C)+T denote the vanilla classifier and AE-C trained with temperature T of 2.5, respectively. C+GCE and (AE-C)+GCE denote the vanilla classfier and AE-C where trained with GCE objective, respectively. Note that FGSM attacks were conducted in (a), (c), (e) and (g), and PGD attacks were conducted in (b), (d), (f) and (h).}
	\label{fig:2}
	\vspace{-0.3cm}
\end{figure*}

To verify that the achieved adversarial accuracy using the autoencoder does not rely on the gradient masking effect, we examined in various ways. Firstly, adversarial defenses using gradient masking effects are often more vulnerable to black-box attacks~\cite{athalye2018obfuscated}. Thus, we conducted black-box attacks against the vanilla classifier and the AE-C using FGSM and PGD. We generated adversarial examples using two types of substitute classifiers \cite{papernot2017practical} for a fair comparison. 
First, we adopt different classifier network than the vanilla network to generate adversarial examples. The autoencoder used the same structure as the vanilla network. Specifically, we used LeNet-like model for MNIST and Resnet-110 \cite{he2016deep} for CIFAR10 as substitute classifiers. Also, since the autoencoder is trained to be unable to reconstruct perfectly using dimensionality reduction \cite{goodfellow2016deep}, concerns about the diminishing effect of the adversarial perturbations may be raised. Therefore, we also considered the second type. For the second type, we adopt the same structured network with the vanilla classifiers and AE-C as substitute classifiers. Table~\ref{tab:2} (a) and (b) indicate the above-mentioned first and second cases, respectively. Table~\ref{tab:2} shows that the autoencoder is also robust in the black-box environment.

Secondly, we examined the effect of the autoencoder on other defense methods. We adopt two gradient masking based approaches; using temperature $\mathit{T}$ of 2.5 and Guided Complement Entropy (GCE) \cite{chen2019improving}. GCE aims to flatten the output weights of false classes. Because GCE uses a kind of regularization, and we observe black-box attacks successfully defeat this method, we categorized this method to one of the gradient masking based methods.

Following the first row of Fig.~\ref{fig:2}, in the white-box environment, the experiment shows that the adversarial accuracy of the AE-C with the gradient masking is higher than that of using the gradient masking alone. We also tested in the black-box environment as shown in the second row of Fig.~\ref{fig:2}. Using temperature or the GCE alone was not robust against black-box attacks at all, as expected. On the other hand, adversarial accuracy increased when used with an autoencoder. These results also show that the autoencoder has a different effect from the gradient masking effect. 

\vspace{-0.3cm}
\subsection{Robust Features of Autoencoder}
\label{sec:typestyle}

\begin{figure}[t]
	\begin{minipage}[b]{1.0\linewidth}
		\centering
		\centerline{\includegraphics[width=7.7cm]{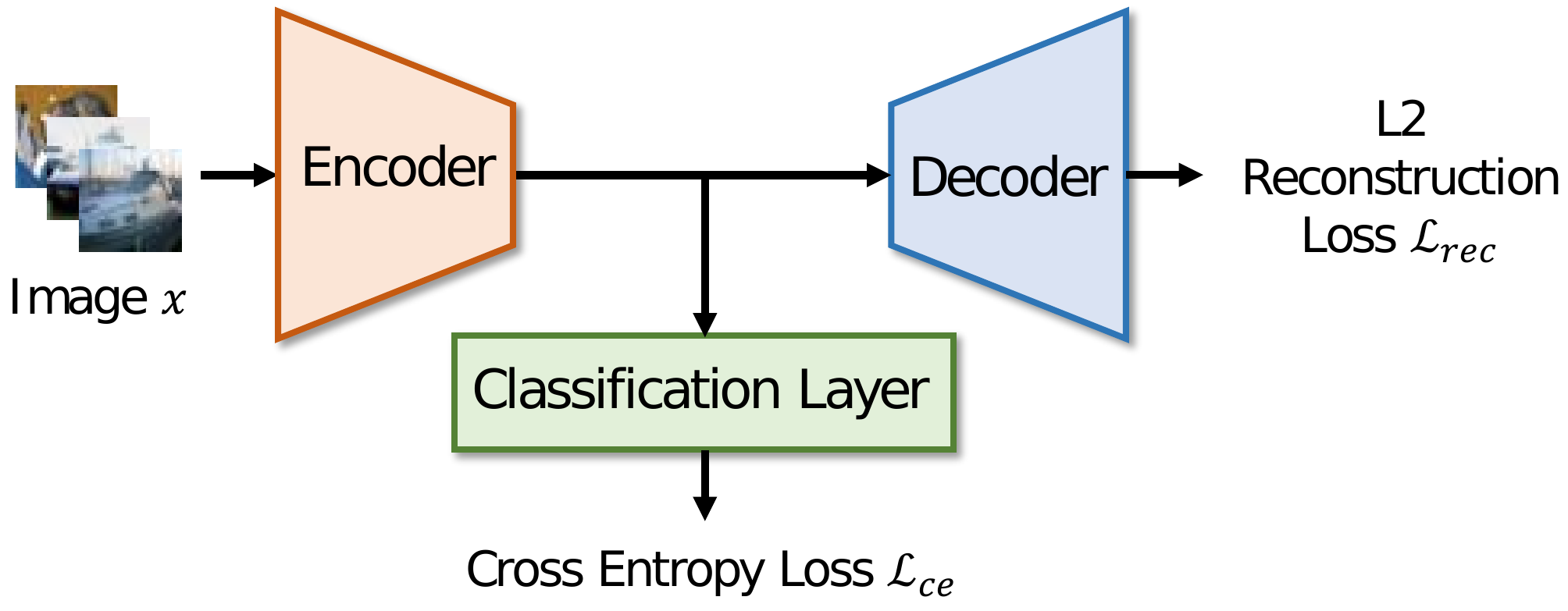}}
	\end{minipage}
	\vspace{-0.3cm}
	\caption{Experiment to see the effect of autoencoder on the features used by the classifier.}
	\label{fig:3}
	\vspace{-0.5cm}
\end{figure}\vspace{-0.3cm}

Recently, some studies~\cite{tsipras2018robustness, ilyas2019adversarial} argue that to obtain adversarial robustness, DNNs should learn robust features which are robust to adversarial perturbations. They pointed out that the adversarial vulnerability may stem from non-robust features that DNNs learn to maximize classification accuracies. We hypothesized that the autoencoder uses robust features for reconstruction so that the autoencoder is somewhat robust against adversarial attacks. To verify this hypothesis, we conducted an experiment by constructing a network as shown in Fig.~\ref{fig:3}. Classification is performed on latent features that have passed through the encoder. The encoder composed of Convolutional Neural Networks (CNNs) roles a feature extractor and the classification layer classifies the encoded features. The network was jointly trained to minimize L2 reconstruction loss of the autoencoder and cross-entropy loss of the classifier. By training the two losses together, we expected the features used by the classifier to be affected by the autoencoder. We conducted the experiments on MNIST dataset to control the optimization accurately. We planned two different training tasks to control the effect of the autoencoder. Firstly, given the cross-entropy loss $\mathcal{L}_{\mathrm{ce}}$ and the L2 reconstruction loss $\mathcal{L}_{\mathrm{rec}}$, we regulated the effect of the autoencoder by a parameter $\lambda$. Total loss $\mathcal{L}$ becomes as follows:\vspace{-0.5cm}

\begin{align}
\mathcal{L} = \mathcal{L}_{\mathrm{ce}}+\lambda\mathcal{L}_{\mathrm{rec}}.
\label{eq:reg}
\end{align}
We optimized our network to minimize the total loss $\mathcal{L}$. We will call this method as a regularization method.

\begin{table}[]
	\centering
	\caption{Effect of an autoencoder with respect to $\lambda$ and $\gamma$. The values of $\varepsilon$ are the values when the image is normalized from 0 to 255. (a) Regularization method. (b) Alternative method.}\vspace{-0.3cm}
	\begin{subtable}{.48\textwidth}\centering
		\begin{tabular}{c|c|ccc}
			& $\varepsilon$  & $\lambda$=0 & $\lambda$=1 & $\lambda$=4 \\ \hline
			\multirow{2}{*}{FGSM} & 20 & 63.73 & 69.07 & \textbf{72.48} \\
			& 50 & 9.3   & 11.96 & \textbf{13.52} \\ \hline
			\multirow{2}{*}{PGD}  & 20 & 28.78 & 36.53 & \textbf{39.86} \\
			& 50 & 0.02  & 0.20  & \textbf{0.51} 
		\end{tabular}\label{tab:3a}\caption{}
	\end{subtable}\vspace{-0.2cm}
	\begin{subtable}{.48\textwidth}\centering
		\begin{tabular}{c|c|ccc}
			& $\varepsilon$  & $\gamma$=0 & $\gamma$=1 & $\gamma$=2 \\ \hline
			\multirow{2}{*}{FGSM} & 20 & 63.73 & 69.02 & \textbf{70.26} \\
			& 50 & 9.3   & 13.60 & \textbf{14.67} \\ \hline
			\multirow{2}{*}{PGD}  & 20 & 28.78 & 40.40 & \textbf{49.51} \\
			& 50 & 0.02  & 0.37  & \textbf{4.89} 
		\end{tabular}\label{tab:3b}\caption{}\vspace{-0.6cm}
	\end{subtable}\vspace{-0.2cm}
\end{table}
\label{tab:3}

Secondly, in every training step, we minimized $\mathcal{L}_{\mathrm{ce}}$ in advance of $\mathcal{L}_{\mathrm{rec}}$. After that, the network was optimized to minimize $\mathcal{L}_{\mathrm{rec}}$ for $\gamma$ times. By choosing a parameter $\gamma$, similar to $\lambda$, we tried to control the degree of effects of the autoencoder. We will call this as an alternative method. 

As shown in Table 3, it seems that the increased influence of the autoencoder in the feature learning process made the classification more robust to adversarial examples. However, following our observations, $\lambda$ larger than 4 and $\gamma$ larger than 2 rather made overall adversarial accuracy decrease. We think this is because the priority of the classification, the primary objective of our process, became too lower than the reconstruction.

Why are the features used by autoencoders robust? We assume that the adversarial examples are generated using class-related information, and autoencoder learns features without any class-related information. Thus, the features used by autoencoders can become less sensitive to adversarial attacks. Making the features less discriminative (\textit{i.e.} less class-related) accords well with the well-known trade-off between robustness and accuracy~\cite{madry2017towards, tsipras2018robustness}. There are many observations that the accuracy on clean data decreases when the classifier is adversarially trained~\cite{madry2017towards, zhang2019defense}.

\begin{figure}[t]
	\begin{subfigure}{.23\textwidth}
		\centering
		\includegraphics[width=1\linewidth]{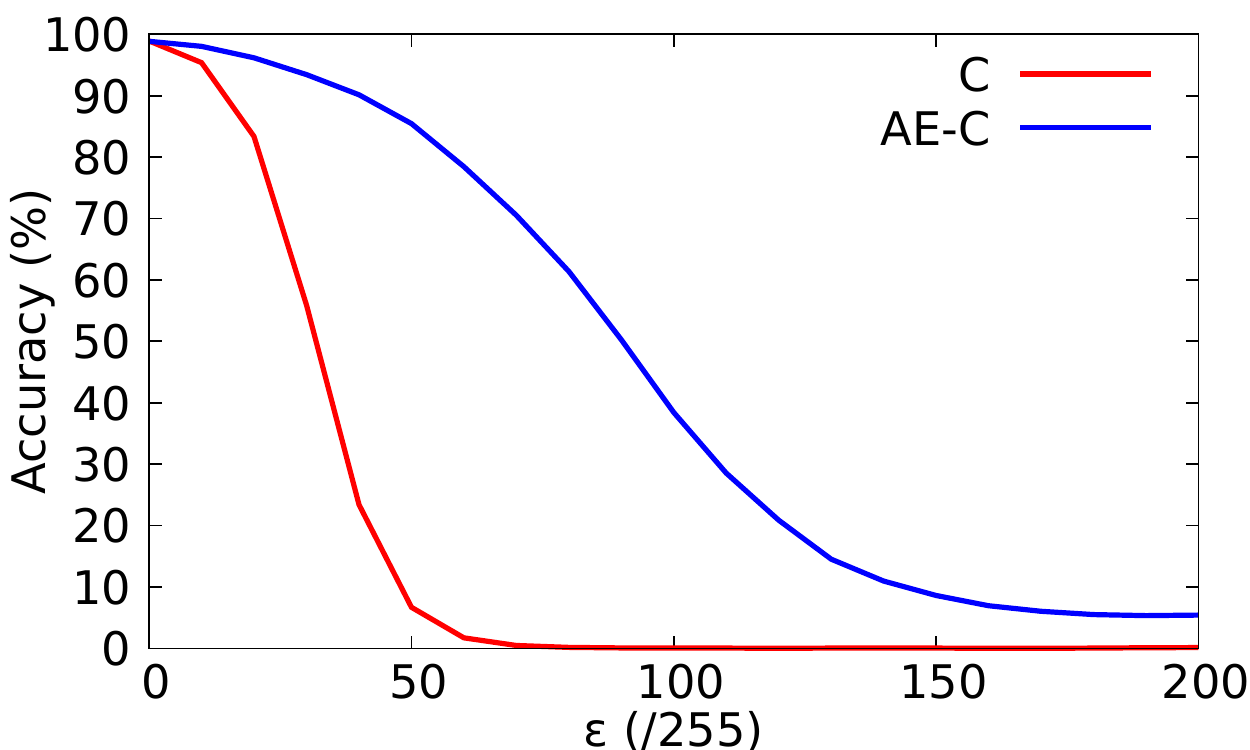}  
		\caption{}
		\label{fig:4a}
	\end{subfigure}\hfill
	\begin{subfigure}{.23\textwidth}
		\centering
		\includegraphics[width=1\linewidth]{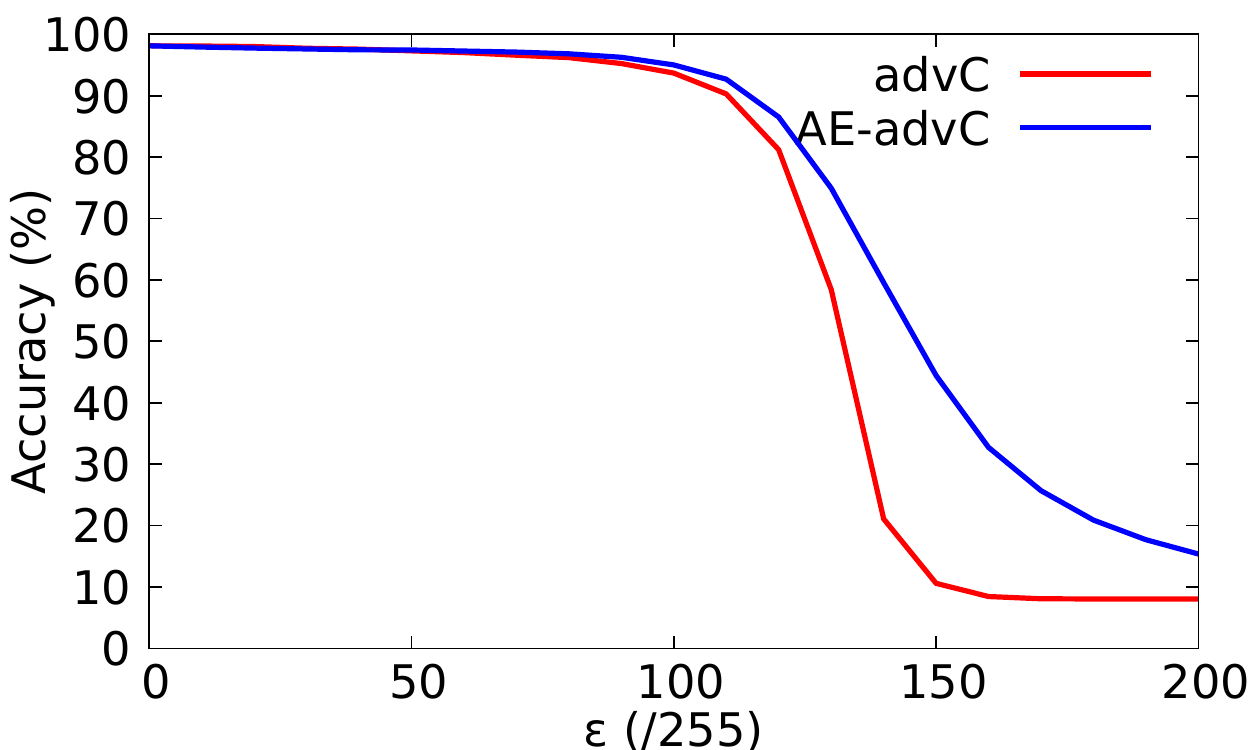}  
		\caption{}
		\label{fig:4b}
	\end{subfigure}\hfill
	\begin{subfigure}{.23\textwidth}
		\centering
		\includegraphics[width=1\linewidth]{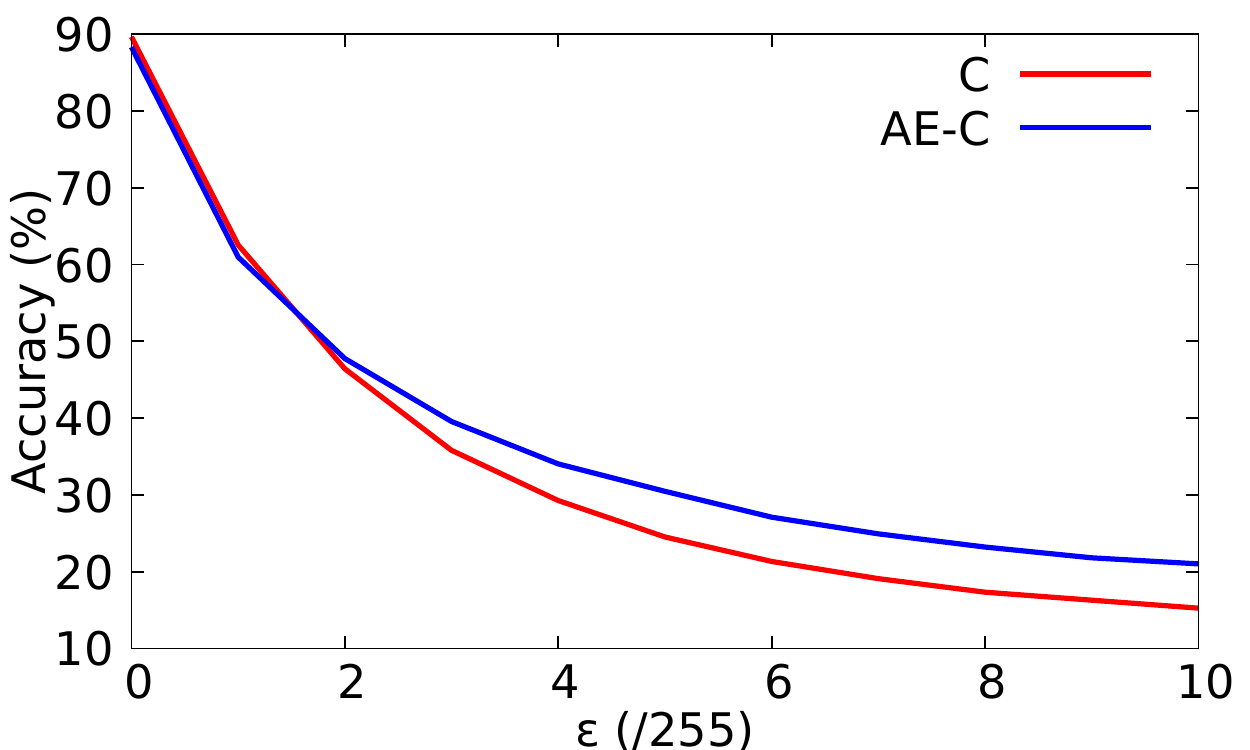}  
		\caption{}
		\label{fig:4c}
	\end{subfigure}\hfill
	\begin{subfigure}{.23\textwidth}
		\centering
		\includegraphics[width=1\linewidth]{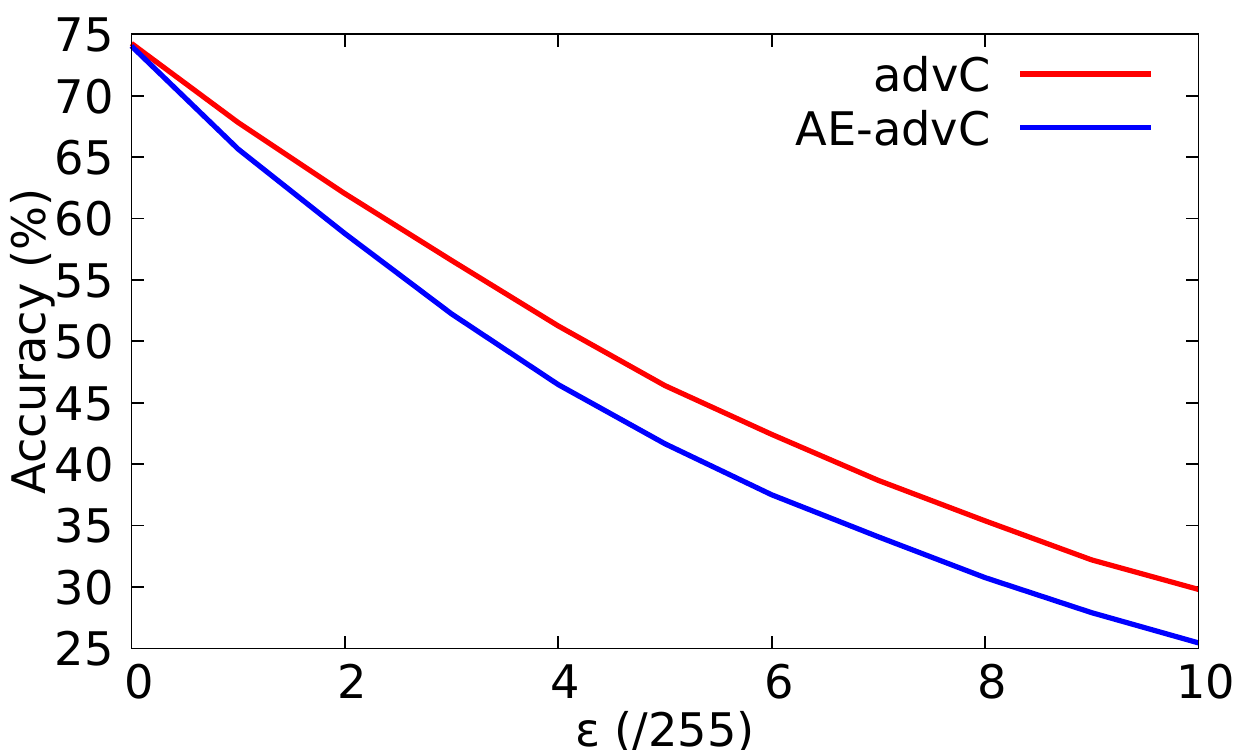}  
		\caption{}
		\label{fig:4d}
	\end{subfigure}
	\vspace{-0.4cm}
	\caption{Combining autoencoder with PGD training. (a) and (b) are conducted on MNIST, and (c) and (d) are conducted on CIFAR10. PGD trained classifiers are denoted as advC.}
	\label{fig:4}
	\vspace{-0.3cm}
\end{figure}

Therefore, we also combined adversarial training with the autoencoder. If the adversarial robustness of the autoencoder and the adversarial training stem from the same cause, the performance of combining the autoencoder and the adversarial training will not be much enhanced than the adversarial training alone. We adopted PGD training~\cite{madry2017towards}. The results are shown in Fig.~\ref{fig:4}. Comparing Fig.~\ref{fig:4} (a) and (b), the accuracy gap became smaller when the adversarially trained classifier was used in the case of MNIST. Nevertheless, adversarial accuracy increased significantly. In the case of CIFAR10, when the autoencoder was attached to the adversarially trained classifier, the adversarial accuracy rather reduced.

To examine the cause of the differences between the two datasets, we compared the amplitudes of the gradients. Reducing gradients is one of the gradient masking approaches which makes it hard to create adversarial examples. We measured the gradients after the softmax layer with respect to the input layer as follows:\vspace{-0.2cm}
\begin{align}
\sum_{i=1}^{N}\left| \nabla f(x_i) \right|_1/N,
\label{eq:gradient}
\end{align}
where $x_i$ is $i$-th input, $f$ is the whole classifier including softmax layer, and $N$ is total number of test set. 

As shown in Table~\ref{tab:4}, the gradients became higher for CIFAR10 and lower for MNIST when using the autoencoders. For MNIST, the effect of the reduced gradients may also have contributed to the adversarial accuracy in the white-box environments. Thus, the effect can be illustrated by the fact that the accuracy gap in the black-box is smaller than in the white-box as shown in Table~\ref{tab:1} and~\ref{tab:2}. Specifically, when the vanilla classifer achieved about 70\% accuracy, AE-C achieved about 89\% adversarial accuracy in the white-box environment while it achieved only about 73\% in the black-box environment. Differences between the two datasets in Fig.~\ref{fig:4} can also be caused by increased and decreased gradients.
\begin{table}[t]
	\vspace{-0cm}
	\renewcommand{\tabcolsep}{0.5mm}
	\caption{Mean of gradient amplitude (G)}
	\vspace{-0.4cm}
	\begin{center}
		\resizebox{0.9999\linewidth}{!}
		{
			
			\begin{tabular}{c|cc|cc}
				\textbf{Dataset} & \multicolumn{2}{c|}{\textbf{CIFAR10 (VGG)}} & \multicolumn{2}{c}{\textbf{MNIST (LeNet)}} \\ \hline
				\textbf{Model}   & \textbf{C}       & \textbf{AE-C}      & \textbf{C}     & \textbf{AE-C}     \\ \hline
				\textbf{G}  & $5.9\times10^{-3}$         & $\mathbf{6.9\times10^{-3}}$            & $\mathbf{8.9\times10^{-4}}$       & $4.3\times10^{-3} $        
			\end{tabular}\label{tab:4}\vspace{-0.5cm}
		}
	\end{center}
	\vspace{-8mm}
\end{table}


Following the experimental results, we suggest that the adversarial robustness of the autoencoders may stem from the features used in the autoencoders. There seems to be adversarial robustness more than the effect of reduced gradients. This adversarial robustness may be the same as the robustness gained through the adversarial training.
\vspace{-0.4cm}
\section{Conclusion}
\label{sec:majhead}

For the safe use of the DNNs, the adversarial robustness must be addressed. In this paper, we revisited the role of the autoencoder in adversarial settings. Improved adversarial accuracy was observed in both white and black-box environments. Experimental results showed that the improved adversarial accuracy differs from gradient masking effect. The adversarial robustness may stem from the features that the autoencoder learns. Moreover, the robust features can be the same as the cause of adversarial robustness of adversarial training. The adversarial robustness of the autoencoder could be hints to the unresolved question: “Where comes the adversarial vulnerability?”




\bibliographystyle{IEEEbib}
\bibliography{refs}

\end{document}